\documentclass{svproc}
\usepackage{epsfig}
\usepackage{subcaption}
\usepackage{amsmath}
\usepackage{amssymb}
\usepackage{bm}
\newcommand{\comment}[1]{}
\usepackage{url}

\begin{document}
\mainmatter              
\title{Dynamic Model of Planar Sliding}
\titlerunning{Dynamic Sliding}  
%
\author{Jiayin Xie\inst{1} \and Nilanjan Chakraborty\inst{2}
}
\authorrunning{J. Xie and N. Chakraborty} 
%
%
\institute{Stony Brook University, Stony Brook NY 11733, USA,\\
\email{jiayin.xie@stonybrook.edu},
\and
Stony Brook University, Stony Brook NY 11733, USA,\\ \email{nilanjan.chakraborty@stonybrook.edu}}

\maketitle              

\begin{abstract}
In this paper, we present a principled method to model general planar sliding motion with distributed patch contact between two objects undergoing relative sliding motion. The effect of contact patch can be equivalently modeled as the contact wrench at one point contact. We call this point {\em equivalent contact point (ECP)}. Our dynamic model embeds ECP within the Newton-Euler equations of slider's motion and friction model. 
The discrete-time motion model that we derive consists of a system of quadratic equations relating the contact wrench and slip speed. This discrete-time dynamic model allows us to solve for the two components of tangential friction impulses, the friction moment, and the slip speed. The state of the slider as well as the ECP can be computed by solving a system of linear equations once the contact impulses are computed. In addition, we derive the closed form solutions for the state of slider for quasi-static motion. Furthermore, in pure translation, based on the discrete-time model, we present the closed form expressions for the friction impulses that acts on the slider and the state of the slider at each time step. Our results are dependent on the rigid body assumption and a generalized Coulomb friction model, which assumes that the contact force and moment lies within a convex cone and the friction force is independent of contact area. The results are not dependent on the exact knowledge of contact geometry or pressure distribution on the contact patch. Simulation examples are shown with both convex and non-convex contact patches to demonstrate the validity of our approach.  
\end{abstract}
\section{Introduction}
In robotic manipulation, a key problem is how to move an object from one configuration to another. There are two ways of manipulating objects, namely, prehensile manipulation and non-prehensile manipulation. In prehensile manipulation, the robot grasps the object and moves it so that all the external wrenches acting on the object through manipulator or gripper during the motion is resisted. In non-prehensile manipulation one manipulates an object without grasping the object. Examples of non-prehensile manipulation includes throwing~\cite{mason1993dynamic,huang1995impulsive,zhu1996releasing}, batting~\cite{andersson1988robot,liu2012racket}, pushing~\cite{lynch1996stable,lynch1999locally,lynch1992manipulation,mason1986mechanics} and vibratory motion~\cite{vose2007vibration,vose2009friction}.For non-prehensile manipulation, where the object being manipulated slides over a support surface, a key aspect to designing planning and control algorithms is the ability to predict the motion of the sliding object. In this paper, our goal is to study the problem of motion prediction of an object sliding on a support surface.

A key aspect of planar sliding motion is that there is usually a non-point patch contact between the slider (or sliding object) and the support. The state of the slider depends on the applied external forces  and the friction forces which distribute over the contact patch.
 The effect of the contact patch can be modeled equivalently by the sum of the total distributed normal and tangential force acting at one point and the net moment about this point due to the tangential contact forces. This point is called the {\em center of friction} in~\cite{mason1986mechanics}. If we assume that the motion of the slider is quasi-static (i.e., the inertial forces are negligible and thus the friction forces balance the applied forces) and center of mass lies on the support plane, the center-of-friction  directly coincide with the center of mass, and closed-form expressions can be developed for motion prediction~\cite{lynch1996stable,lynch1992manipulation,mason1986mechanics}. However, for dynamic sliding, when the inertial forces cannot be neglected and center of gravity is above the support plane, the center of friction can vary within the convex hull of the contact patch, and there is no method in the literature for computing it.  

 The existing approach is to use a dynamic simulation algorithm  with contact patch usually approximated with three support points (chosen in ad-hoc manner). The reason for choosing three support points is that most dynamic simulation algorithms that are usually used for motion prediction implicitly assumes a point contact model and choosing $3$ support points ensure that the force distribution at the three points is unique (if four or more points are used, the force distribution will not be unique for the same equivalent force and moment acting on the object). If the center of friction lies within the convex hull of the chosen support points, the motion predicted will be accurate. However, if the center of friction is outside the convex hull, then the predicted motion will not be accurate. Furthermore, since we do not know the center of friction, we will not know when the predicted motion is inaccurate. Note that the accuracy issue arises here due to the {\em ad hoc} approximation of the patch contact and not due to other sources of inaccuracy like contact friction model or model parameters.

In this paper, based on our previous work of nonlinear complementarity problem-based dynamic time-stepper with convex contact patch~\cite{xie2016rigid}, we present a dynamic model for sliding, where no {\em ad hoc} approximation is made for the patch contact. We model the patch contact with a point contact, called the {\em equivalent contact point} (ECP). {\em The ECP is defined as a unique point in the contact patch where the net moment due to the normal contact force is zero~\cite{xie2016rigid}}. We show that the computation of the contact forces/impulses and the state of the object and ECP can be decoupled for planar sliding. The contact impulses can be computed by solving four quadratic equations in four variables, namely, the two components of tangential impulse, the frictional moment, and the slip speed. The state variables, namely, the position, orientation, linear, and angular velocities of the object as well as the ECP can be computed by solving a system of linear equations once the contact impulses are computed. {\em Note that the ECP as defined here is the same as the center of friction}. 
The presentation of the decoupled set of quadratic and linear equations for computing the contact impulse, the ECP, and the state of the slider is the key contribution of this paper. We show that closed form solutions for the state of the slider can be derived for quasi-static motion (which is same as those previously obtained in the literature). For pure translation also, closed form solutions can be derived for the contact impulse, the state of the object and the ECP. We also present numerical simulation results comparing the model that we derive to the solution of the NCP model from~\cite{xie2016rigid}.  Our results are dependent on the rigid body assumption and a generalized Coulomb friction model, which assumes that (a) the contact force and moment lies within a convex cone and (b) the friction force is independent of contact area and only dependent on magnitude of normal force. The results are not dependent on the exact knowledge of contact geometry or pressure distribution on the contact patch.

\section{Related Work}
\label{sec:rw}

During sliding, friction plays an important role in determining the motion of object. Coulomb's friction law (also called Amonton, da vinci or dry friction law)~\cite{coulomb1821theorie}, which suggests that the friction force should be proportional to the normal force and opposed to the direction of sliding is a popular friction model. There have been many efforts to extend Coulomb's law into general sliding planar motion~\cite{mason1986mechanics,jellett1872treatise,prescott1929mechanics,macmillan1936dynamics} where one has to consider both force and moment due to the contact. In~\cite{goyal1991planar}, the authors presented a geometric description, which is so called {\em limit surface}, for the relationship between the motion of the slider and the total frictional support force. In~\cite{howe1996practical},  the authors present multiple approximations (square pyramids, cones and ellipsoids, etc.) for the limit surface based on experimental results. 

From basic physics, for patch contact, there exists a unique point on the contact patch where the net moment due to the normal contact force is zero. This point is called the {\em center of friction}. For pure translation, the system of frictional forces arising in the contact patch may be reduced to a single force acting through the center of friction~\cite{macmillan1936dynamics}. In~\cite{mason1986mechanics}, based on the concept of center of friction, the authors develop the {\em voting theorem} to determine whether an object will rotate and in which direction when it is pushed. However, in general, computing the center of friction from it's definition is not possible without knowing the pressure distribution between the two sliding objects. Therefore, previous models for sliding motion~\cite{lynch1996stable,lynch1992manipulation} make assumptions like quasi-static motion, uniform pressure distribution in the contact patch and isotropic friction, so that the center of friction is computable.

There also exists attempts to apply data-driven techniques to the problem of sliding motion. In~\cite{YuB+2016}, the authors record the motion and forces of the slider for different shape and material. They also present the empirical analysis of sliding motion. In~\cite{ZhouMPB18}, the authors develop a data-driven but physics-based method for planar sliding, which approximates the mapping based on limit surface between frictional loads and motion twists. 

In our previous work~\cite{xie2016rigid}, we developed a principled method to model line or surface contact between objects, in which, the effect of contact patch is modeled equivalently as point contact. We called this point {\em equivalent contact point} (ECP). Although this is same as center of friction, it was conceptualized to prevent penetration between contacting objects. We showed that this point can be computed along with contact wrenches if we formulate a non-linear complementarity problem that simultaneously solves the contact detection problem along with the numerical integration of the equations of motion. Note that this is different from the current paradigm of dynamic simulation, where the contact detection and numerical integration of the equations of motion are decoupled and are done in a sequence. Consequently for non-point contact, the contact detection problem is ill-posed, as there are infinitely many points that are valid solutions.  In this paper, based on our previous general model of equations of motion for bodies in intermittent contact, we derive a dynamic model for sliding motion, where the contact patch between slider and ground is equivalently modeled with an ECP. We assume a friction model that is based on maximum power dissipation principle and it assumes all the possible contact forces or moments should lie within an ellipsoid (similar to ~\cite{goyal1991planar}). Note that we do not make any assumptions about the pressure distribution in the contact patch.

\comment{In general, without knowing the external forces and moments (including applied and friction forces), the ECP or center of friction is not computable. In addition, the acceleration of a body whose positions of external forces (contact point of pushing, center of gravity, etc.) are above or below the support plane will, in general, cause the shift in the pressure distribution and corresponding shift in the center of friction or ECP~\cite{mason2001mechanics}. Due to the reasons above,} 

\section{Dynamics of Bodies in Contact}
In this section, we present the general equations of motion of rigid bodies moving with respect to each other with non-point contact. For simplicity of exposition we assume one body to be static. The general equations of motion of the moving body has three key parts (a) Newton-Euler  differential equations of motion giving state update, (b) algebraic and complementarity constraints modeling the fact that two rigid bodies cannot penetrate each other and (c) model of the frictional force and moments acting on the contact patch. Let $\bm{\nu} = [\bm{v}^T ~\bm{\omega}^T]^T$ be the generalized velocity of the rigid body, where $\bm{v} \in \mathbb{R}^3$ is the linear velocity and $\bm{\omega} \in \mathbb{R}^3$ is the angular velocity of the rigid body. Let $\bm{q}$ be the configuration of the rigid body, which is a concatenated vector of the position and a parameterization of the orientation of the rigid body. 

{\bf Newton-Euler Equations of Motion}:
The Newton-Euler equations of motion of the rigid body are:
\begin{equation} \label{eq_general}
\bm{M}(\bm{q})
{\dot{\bm{\nu}}} = 
\bm{W_{n}}\lambda_{n}+
\bm{W_{t}} \lambda_{t}+
\bm{W_{o}} \lambda_{o}+
\bm {W_{r}}\lambda_{r}+
\bm{\lambda_{app}}+\bm{\lambda_{vp}}
\end{equation}
where $\bm{M}(\bm{q})$ is the inertia tensor, $\bm{\lambda}_{app}$ is the vector of external forces and moments (including gravity), $\bm{\lambda}_{vp}$ is the  centripetal and Coriolis forces. The magnitude of the normal contact force is $\lambda_n$. The magnitude of tangential contact forces are $\lambda_t$ and $\lambda_o$. The magnitude of the moment due to the tangential contact forces about the contact normal is $\lambda_r$. The vectors $\bm{W}_n$, $\bm{W}_t$, $\bm{W}_o$ and $\bm{W}_r$ map the contact forces and moments from the contact point to the center of mass of the robot. The expressions of $\bm{W}_n$, $\bm{W}_t$, $\bm{W}_o$ and $\bm{W}_r$ are:
\begin{equation}
\begin{aligned}
\label{equation:wrenches}
\bm{W_{n}} =  \left [ \begin{matrix} 
\bm{n}\\
\bm{r}\times \bm{n}
\end{matrix}\right],
\quad 
\bm{W_{t}} =  \left [ \begin{matrix} 
\bm{t}\\
\bm{r}\times \bm{t}
\end{matrix}\right],
\quad
\bm{W_{o}} =  \left [ \begin{matrix} 
\bm{o}\\
\bm{r}\times \bm{o}
\end{matrix}\right],
\quad 
\bm{W_{r}} =  \left [ \begin{matrix} 
\bm{0}\\
\ \ \bm{n} \ \
\end{matrix}\right]
\end{aligned}
\end{equation}
where $(\bm{n},\bm{t},\bm{o}) \in \mathcal{R}^3$ are the axes of the contact frame, $\bm{0} \in \mathcal{R}^3$ is a column vector with each entry equal to zero. The vector $\bm{r} = [a_x-q_x,a_y-q_y,a_z-q_z]$ is the vector from ECP, $\bm{a}$, to center of mass (CM), where $(q_x, q_y, q_z)$ is the position of the CM. Please note that Equation~\eqref{eq_general} is a system of $6$ differential equations.

{\bf Modeling Rigid body Contact Constraints}:
The contact model that we use is a complementarity-based contact model as described in~\cite{xie2016rigid,chakraborty2014geometrically}. In~\cite{xie2016rigid}, we introduced the notion of an equivalent contact point (ECP) to model non-point contact between objects. Equivalent Contact Point (ECP) is a unique point on the contact surface that can be used to model the surface (line) contact as point contact where the integral of the total moment (about the point) due to the distributed normal force on the contact patch is zero. The ECP defined here is the same as the center of friction. However, we believe that ECP is an apt name, because it allows us to enforce constraints of non-penetration between two rigid bodies. For the special case of planar sliding motion, since there is always contact, we do not need to write down the equations coming from the collision detection constraints as done in~\cite{chakraborty2014geometrically,xie2016rigid} for computing the ECP. These constraints are trivially satisfied. However, we do need to use the ECP in the equations of motion as we do in the later sections. 

{\bf Friction Model}:
We use a friction model based on the maximum power dissipation principle that generalizes Coulomb's friction law. The maximum power dissipation principle states that among all the possible contact forces and moments that lie within the friction ellipsoid, the forces that maximize the power dissipation in the contact patch are selected. Mathematically, 
\begin{equation}
\begin{aligned}
\label{equation:friction}
{\rm max} \quad -(v_t \lambda_t + v_o\lambda_o + v_r \lambda_r)\\
{\rm s.t.} \quad \left(\frac{\lambda_t}{e_t}\right)^2 + \left(\frac{\lambda_o}{e_o}\right)^2+\left(\frac{\lambda_r}{e_r}\right)^2 - \mu^2 \lambda_n^2 \le 0
\end{aligned}
\end{equation}
where $\lambda_t$, $\lambda_o$, and $\lambda_r$ are the optimization variables. The parameters, $e_t$, $e_o$, and $e_r$ are positive constants defining the friction ellipsoid and $\mu$ is the coefficient of friction at the contact~\cite{howe1996practical,trinkle1997dynamic}.
We use the contact wrench at ECP to model the effect of entire distributed contact patch. Therefore $v_t$ and $v_o$ are the tangential components of velocity at ECP; $v_r$ is the relative angular velocity about the normal at ECP.  
Note that, the ellipsoid constraint in our friction model is the constraint on the friction force and moment that acts on the slider during the motion. This friction model has been previously proposed in the literature~\cite{goyal1991planar} and has some experimental justification~\cite{howe1996practical}. There is {\em no assumption made on the nature of the pressure distribution between the two surfaces}.  

Using the Fritz-John optimality conditions of Equation~\eqref{equation:friction}, we can write~\cite{trinkle2001dynamic}:
\begin{align}
\label{eq:friction_1}
0&=
e^{2}_{t}\mu \lambda_{n} 
v_t+
\lambda_{t}\sigma\\
\label{eq:friction_2}
0&=
e^{2}_{o}\mu \lambda_{n}  
v_o+\lambda_{o}\sigma\\
\label{eq:friction_3}
0&=
e^{2}_{r}\mu \lambda_{n}v_r+\lambda_{r}\sigma\\
\label{eq:complement_friction}
0& \le \mu^2\lambda_n^2- \lambda_{t}^2/e^{2}_{t}- \lambda_{o}^2/e^{2}_{o}- \lambda_{r}^2/e^{2}_{r} \perp \sigma \ge 0
\end{align}
where $\sigma$ is a Lagrange multiplier corresponding to the inequality constraint in~\eqref{equation:friction}. 


\section{Equations of motion for planar sliding}
\label{sec:Mp}
The dynamic model presented in the previous section is a general model for an object moving on a planar surface with intermittent contact (that can be non-point) between the object and the surface. In this section, we will assume that the motion between the two objects is planar sliding and derive a simpler set of equations that are valid for planar sliding.

\begin{figure}
\centering
fi\includegraphics[width=2in]{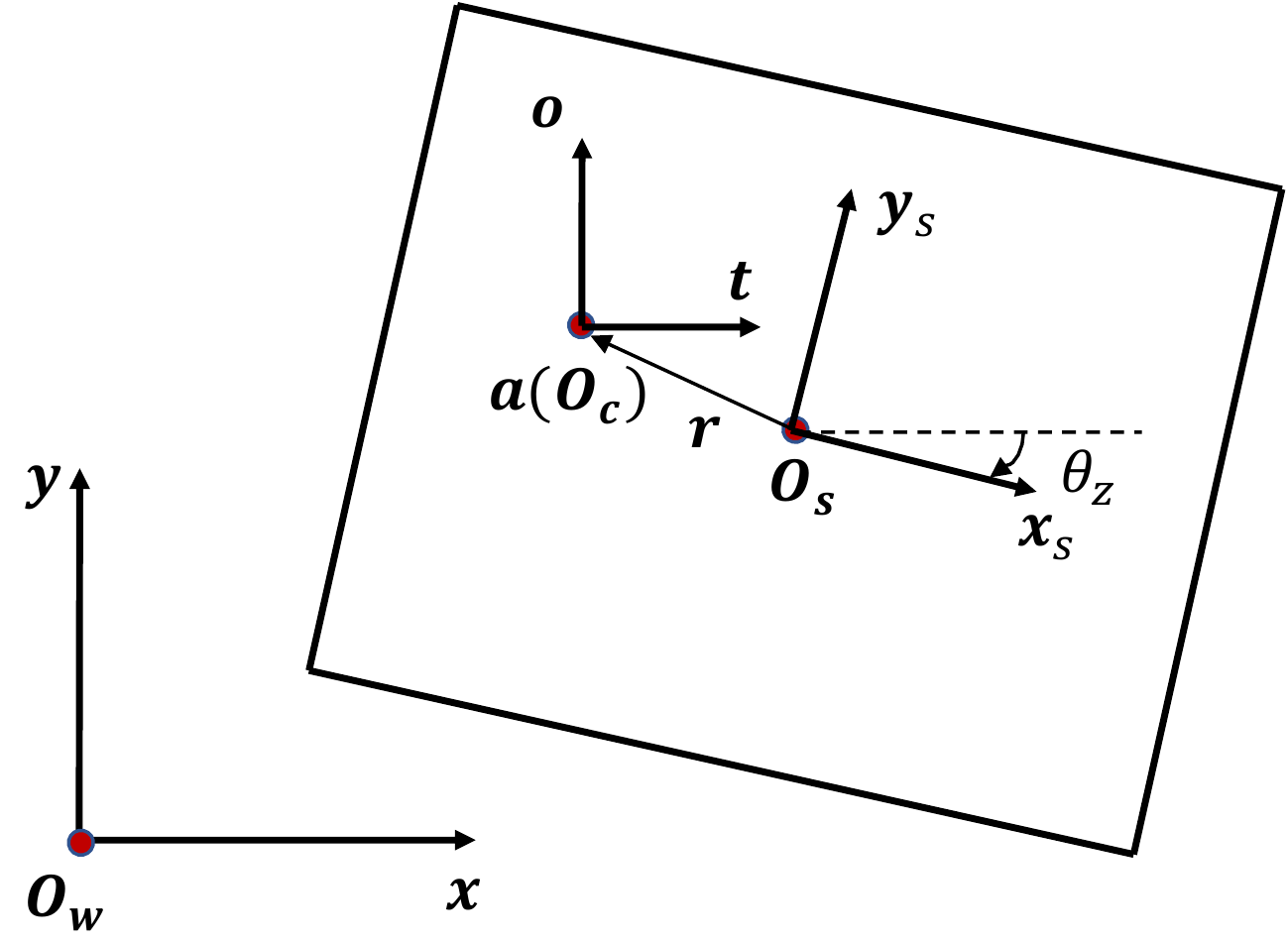}
\caption{Slider with square contact patch on the horizontal support plane.}
\label{figure_1} 
\end{figure}

Figure~\ref{figure_1} shows a schematic sketch of a slider (assumed to be a rigid body) that has planar surface contact with the support surface. We assume that the motion of the slider is planar, i.e., the slider can rotate and translate along the planar support surface but cannot topple or lose contact with the support surface. Let $\mathcal{F}_w$ with origin $\bm{O}_w$ be the world frame fixed on the support surface. Let $\mathcal{F}_s$ with origin $\bm{O}_s$ be the slider frame attached to the slider's center of mass (CM).  Note that the coordinates of the CM in the world frame, $\mathcal{F}_w$ is ($q_x$, $q_y$, $q_z$).
Since, the slider undergoes planar motion,
the configuration of the slider is $\bm{q} = [q_x,q_y,\theta_z]$, where $\theta_z$ is the orientation of $\mathcal{F}_s$ relative to $\mathcal{F}_w$. Let $\mathcal{F}_c$ with origin $\bm{O}_c$ be the contact frame. The origin $\bm{O}_c$ is the equivalent contact point (ECP) of the contact patch and we denote the position of $\bm{O}_c$ in  the world frame, $\mathcal{F}_w$, by $\bm{a}$. The axes of $\mathcal{F}_c$ are chosen to be parallel to $\mathcal{F}_w$. The generalized velocity of the slider is $\bm{\nu} = [v_x,v_y,w_z]$, where $v_x$ and $v_y$ are the $x$ and $y$ components of the velocity of the center of mass, $\bm{O}_s$,  and $w_z$ is the angular velocity about the $z$-axis (normal to the plane of the motion).

\label{subsec:pl}

The external forces acting on the slider includes applied force, gravity force, normal or support force and frictional force. \comment{In this paper, for simplifying exposition, we assume that the applied force is horizontal and the moment it produces is only about the $z$-axis, i.e., the normal applied force $\lambda_z = 0$, and horizontal applied moments $\lambda_{x\tau} = \lambda_{y\tau} =0$. }The generalized applied force is $\bm{\lambda}_{app} = [\lambda_x, \lambda_y,\lambda_z,\lambda_{x\tau},\lambda_{y\tau},  \lambda_{z\tau}]$, which includes tangential and normal forces and moments. The gravitational force is assumed to act at the CM. 

\subsection{Newton-Euler Equations for planar sliding}
\label{subse:eqsm}

For planar motion, the inertia tensor is $\bm{M}(\bm{q}) = diag(m,m,I_z)$, where $m$ is the mass of the slider and $I_z$ represents the moment of inertia about $z$-axis. As mentioned in Section~\ref{subsec:pl}, the configuration of the slider is $\bm{q} = [q_x,q_y,\theta_z]$, 
and the generalized velocity is $\bm{\nu}  = [v_x,v_y,w_z]$. The generalized applied force is $\bm{\lambda}_{app} = [\lambda_x, \lambda_y,\lambda_z,\lambda_{x\tau},\lambda_{y\tau},  \lambda_{z\tau}]$. Without loss of generality, we let unit vectors of the contact frame to be $\bm{n} = [0,0,1], \bm{t} = [1,0,0], \bm{o} = [0,1,0]$. Consequently, $\bm{W}_n$, $\bm{W}_t$, $\bm{W}_o$ and $\bm{W}_r$ (Equation~\eqref{equation:wrenches}) could be written as:
\begin{equation}
\begin{aligned}
\label{equation:wrenches_sim}
\bm{W_{n}} =  \left [ \begin{matrix} 
0\\
0\\
0
\end{matrix}\right],
\quad 
\bm{W_{t}} =  \left [ \begin{matrix} 
1\\
0\\
-(a_y-q_y)
\end{matrix}\right],
\quad 
\bm{W_{o}} =  \left [ \begin{matrix} 
0\\
1\\
a_x-q_x
\end{matrix}\right],
\quad 
\bm{W_{r}} =  \left [ \begin{matrix} 
0\\
0\\
1
\end{matrix}\right]
\end{aligned}
\end{equation}
\noindent
Using~\eqref{equation:wrenches_sim}, and the discussion above, the first, second, and sixth equation in~\eqref{eq_general}, can be written as:
\begin{align}
\label{eq:planar_1}
m&\dot{v}_x= \lambda_t +\lambda_x\\
\label{eq:planar_2}
m&\dot{v}_y = \lambda_o +\lambda_y\\
\label{eq:planar_3}
I_z&\dot{w}_z = \lambda_r +\lambda_{z\tau} -\lambda_t(a_y-q_y)+\lambda_o(a_x-q_x)
\end{align}
\comment{
\begin{align}
\label{eq:planar_1}
m&(v_x^{u+1}-v_x^{u}) = p_t^{u+1} +p_x^{u}\\
\label{eq:planar_2}
m&(v_y^{u+1}-v_y^{u}) = p_o^{u+1} +p_y^u\\
\label{eq:planar_3}
I_z&(w_z^{u+1} - w_z^u) = p_r^{u+1} +p_{z\tau}^u
\end{align}
}
Since, we are assuming that contact is always maintained, the third equation in~\eqref{eq_general} becomes $\lambda_n + \lambda_z - mg =0 $. Furthermore, based on Equations~\eqref{eq:add_1} and~\eqref{eq:add_2} below, we can derive
$[\lambda_{x\tau}(a_x-q_x)+\lambda_{y\tau}(a_y-q_y )]/(q_z-a_z)= -\lambda_t(a_y-q_y)+\lambda_o(a_x-q_x) $.
It makes Equation~\eqref{eq:planar_3} to be:
\begin{equation}
\label{eq:updated}
I_z\dot{w}_z = \lambda_r +\lambda_{z\tau} +[\lambda_{x\tau}(a_x-q_x)+\lambda_{y\tau}(a_y-q_y )]/(q_z-a_z)
\end{equation}

\subsection{Expressions for the ECP}
The fourth and fifth equations in~\eqref{eq_general}, which governs the angular accelerations $\dot{w}_x$ and $\dot{w}_y$, are
\begin{align}
\label{eq:add_1}
I_x\dot{w}_x = \lambda_{x\tau} + \lambda_n(a_y-q_y) - \lambda_o(a_z-q_z) +I_yw_yw_z-I_zw_yw_z \\
\label{eq:add_2}
I_y\dot{w}_y = \lambda_{y\tau} - \lambda_n(a_x-q_x) + \lambda_t(a_z-q_z) -I_xw_xw_z+I_zw_xw_z 
\end{align}
where $I_x$ and $I_y$ are the moment of inertia about the $x$ and $y$ axis respectively. Since the motion is planar, the slider can only rotate about $z$-axis, i.e., $\dot{w_x} = \dot{w_y} = 0$. Equations~\ref{eq:add_1} and~\ref{eq:add_2} provide us the expressions for ECP:
\begin{align}
\label{eq:ECP_1}
a_x &= (\lambda_{y\tau}-\lambda_tq_z)/\lambda_n+q_x\\
\label{eq:ECP_2}
a_y &= (-\lambda_{x\tau}-\lambda_oq_z)/\lambda_n+q_y
\end{align}
Equations~\eqref{eq:ECP_1} and~\eqref{eq:ECP_2} provide us a  expressions for ECP based on the friction forces. From the equations one can deduce that {\em when $q_z = 0$, the ECP ($a_x, a_y$) would be just beneath the CM, i.e., ($q_x, q_y$). When  $q_z > 0$, i.e., CM is above the support plane, ECP may shift from the projection of the CM on the plane}. Furthermore, the variation of tangential friction forces ($\lambda_t$ and $\lambda_o$) or applied moments ($\lambda_{x\tau}$ and $\lambda_{y\tau}$) would cause the shift of ECP during the motion. 

Note that since we have assumed no toppling, we always get a solution for the ECP. However, we can also use the computed ECP to check whether the assumption of no toppling is valid. If the ECP lies outside the convex hull of the contact region between the two objects, the sliding assumption is no longer valid. This can be used even in the discrete time setting to verify that there is no toppling.

\subsection{Friction Model}
For planar sliding, the friction force and moment has to be at the boundary of the friction ellipsoid. Thus, $\sigma > 0$ in the complementarity Equation~\eqref{eq:complement_friction}.
Furthermore, the tangential velocity at ECP is $[v_t, v_o]^T = \bm{v}+ \bm{w}\times\bm{r}$, and $v_r = w_z$. Thus:
\begin{equation}
\label{eq:v_t}
v_t = v_x -w_z(a_y -q_y), \ \ 
v_o = v_y + w_z(a_x - q_x), \ \
v_r = w_z.
\end{equation}
Using Equation~\eqref{eq:v_t} together with the fact that $\sigma > 0$, we can rewrite Equations~\eqref{eq:friction_1} to~\eqref{eq:complement_friction} as
\begin{align}
\label{eq:dy_1}
&0=\mu \lambda_ne_t^2[v_x - w_z(a_y - q_y)] + \lambda_t\sigma  \\
\label{eq:dy_2}
&0=\mu \lambda_ne_o^2[v_y + w_z(a_x - q_x)] + \lambda_o\sigma  \\
\label{eq:dy_3}
&0=\mu \lambda_ne_r^2w_z+ \lambda_r\sigma \\
\label{eq:dy_4}
&0=\mu^2\lambda_n^2 - \lambda_r^2/e_r^2 - \lambda_t^2/e_t^2 - \lambda_o^2/e_o^2 
\end{align}



\subsection{Continuous Time Dynamic Model for Planar Sliding}

The complete continuous time equations of motion for planar sliding are given by (a) the Newton-Euler equations, (Equations~\eqref{eq:planar_1},~\eqref{eq:planar_2} and~\eqref{eq:updated}) (b) the expression for ECP (Equations~\eqref{eq:ECP_1} and~\eqref{eq:ECP_2} ) and (c) the friction model (Equations~\eqref{eq:dy_1} to \eqref{eq:dy_4}). Note that the kinematic map $\dot{\bm{q}} = {\bm \nu}$ is also required.

\subsection{Discrete-time dynamic model}
We use backward Euler time-stepping scheme to discretize the continuous equations of planar sliding motion. Let $t_u$ denote the current time and $h$ be the duration of the time step, the superscript $u$ represents the beginning of the current time and the superscript $u+1$ represents the end of the current time. 
Let $\dot{\bm{\nu}} \approx ( {\bm{\nu}}^{u+1} -{\bm{\nu}}^{u} )/h$ and the impulse $p_{(.)} = h\lambda_{(.)}$. Equations~\eqref{eq:planar_1},~\eqref{eq:planar_2} and~\eqref{eq:updated} become
\begin{align}
\label{eq:planar_1_d}
m&(v_x^{u+1}-v_x^{u}) = p_t^{u+1} +p_x^{u}\\
\label{eq:planar_2_d}
m&(v_y^{u+1}-v_y^{u}) = p_o^{u+1} +p_y^u\\
\label{eq:planar_3_d}
I_z&(w_z^{u+1} - w_z^u) = p_r^{u+1} +p_{z\tau}^u
\end{align}

\comment{
Let $\dot{\bm{q}} \approx ( {\bm{q}}^{u+1} -{\bm{q}}^{u} )/h$, we discretize the Kinematic map (Equation~\eqref{eq:kinematic_map}) and get:
\begin{align}
\label{kinetic_map_d}
q_x^{u+1} = q_x^u + hv_x^{u+1}\\
q_y^{u+1} = q_y^u + hv_y^{u+1}
\end{align}
}
Using Equations~\eqref{eq:planar_1_d} to~\eqref{eq:planar_3_d} and the backward Euler discretization, we can rewrite Equations~Equations~\eqref{eq:dy_1} to~\eqref{eq:dy_4} as
\begin{align}
\label{eq:dy_ts_1}
&0=\mu p_ne_t^2\left(v_x^u + \frac{p_t^{u+1} + p_x^u}{m} + \frac{(p^u_{x\tau}+p_o^{u+1}q_z)[w_z^u + (p_r^{u+1} + p_{z\tau}^u)/I_z]}{p_n}\right) + p_t^{u+1}\sigma^{u+1}  \\
\label{eq:dy_ts_2}
&0=\mu p_ne_o^2\left(v_y^u + \frac{p_o^{u+1} + p_y^u}{m} + \frac{(p^u_{y\tau}-p_t^{u+1}q_z)[w_z^u + (p_r^{u+1} + p_{z\tau}^u)/I_z]}{p_n}\right) + p_o^{u+1}\sigma^{u+1}  \\
\label{eq:dy_ts_3}
&0=\mu p_ne_r^2[w_z^u + (p_r^{u+1} + p_{z\tau}^u)/I_z ] + p_r^{u+1}\sigma^{u+1} \\
\label{eq:dy_ts_4}
&0=\mu^2p_n^2 - (p_r^{u+1}/e_r)^2 - (p_t^{u+1}/e_t)^2 - (p_o^{u+1}/e_o)^2
\end{align}
The equations~\eqref{eq:dy_ts_1} to \eqref{eq:dy_ts_4} is a system of four quadratic equations in the $4$ unknowns, $p_t, p_o, p_r$, and $\sigma$ at the end of the time-step (i.e., with the superscript $u+1$). After solving these system of equations we can obtain the velocities at the end of the time step, $v_x$, $v_y$, $\omega_z$ from the linear equations~\eqref{eq:planar_1_d} to~\eqref{eq:planar_3_d}. The ECPs can be found from Equations~\eqref{eq:ECP_1} and ~\eqref{eq:ECP_2}. {\em Thus, the solution of the dynamic time-stepping problem essentially reduces to the solution of $4$ quadratic equations in $4$ variables}. 



\section{Closed Form Equations For Planar Sliding Motion}
In this section, we study some special cases of planar sliding motion, where we can obtain a closed form solution for the motion as well as the contact wrenches. The two special cases are that of quasi-static sliding, where we know the velocity of the contact point between the slider and pusher and pure translation.

{\bf Quasi-static sliding motion:}
\label{subse:qssm}
In quasi-static sliding, the inertial force can be neglected and the frictional forces dominate the motion of the slider. We assume that the quasi-static sliding is due to an applied force with components $\lambda_x$ and $\lambda_y$ acting on the boundary of the slider at position $(x_c,y_c)$. The associated applied torque about the $z$-axis is $\lambda_{z\tau}$. Based on the equations of sliding motion (Equations~\eqref{eq:planar_1} to~\eqref{eq:planar_3}), the quasi-static motion assumption implies that the friction force and applied force should balance with each other (i.e., $\lambda_t = -\lambda_x$, $\lambda_o = -\lambda_y$, $\lambda_r = -\lambda_{z\tau}$). We take $v_{cx}, v_{cy}$, the velocity components at $(x_c,y_c)$ as the input. This basically says that the point of application of the force can vary during the motion. Thus, the motion of the slider depends on ($v_{cx}, v_{cy}$). Now,
\begin{equation}
\label{eq:cp1}
v_x = v_{cx} +w_z(y_c-q_y), \quad
v_y = v_{cy} -w_z(x_c-q_x).
\end{equation}

Furthermore, the friction moment about normal axis balances with the applied moment and it can be defined by the components of friction force:
\begin{equation}
\lambda_r = (x_c - q_x)\lambda_o - (y_c - q_y)\lambda_t
\end{equation}

Quasi-static model assumes that the ECP or center of friction is just beneath the CM, i.e., $a_x = q_x$, $a_y = q_y$. In addition, the model assumes isotropic friction, which implies $e_t = e_o$. We define the parameter $c = e_r/e_t$. From Equation~\eqref{eq:dy_1} to~\eqref{eq:dy_3}, we get:
\begin{equation}
\label{eq:cp2}
\frac{v_x}{w_z} = c^2\frac{\lambda_t}{\lambda_r}, \quad
\frac{v_y}{w_z} = c^2\frac{\lambda_o}{\lambda_r}
\end{equation}
From the above discussion, using Equations~\eqref{eq:cp1} to~\eqref{eq:cp2}, we can get the closed form expressions for the velocity of the slider ($v_x, v_y, w_z$):
\begin{align}
\label{quasi_1}
v_x &= \frac{[c^2+(x_c-q_x)^2]v_{cx}+(x_c-q_x)(y_c-q_y)v_{cy}}{c^2+(x_c-q_x)^2+(y_c-q_y)^2}\\
\label{quasi_2}
v_y &= \frac{[c^2+(y_c-q_y)^2]v_{cy}+(x_c-q_x)(y_c-q_y)v_{cx}}{c^2+(x_c-q_x)^2+(y_c-q_y)^2}\\
\label{quasi_3}
w_z &= \frac{(x_c-q_x)v_y - (y_c-q_y)v_x}{c^2}
\end{align}

In~\cite{lynch1992manipulation}, the authors also present the closed form solutions for computing the velocity of the slider with quasi-static motion. Note that, if we assume the origin at the CM, i.e., $q_x = q_y = 0$, the Equations~\eqref{quasi_1} to~\eqref{quasi_2} would be equivalent to the equations of quasi-static motion in~\cite{lynch1992manipulation}. 

{\bf Pure translation:}
In this subsection, we derive the closed form expression for pure translation. During pure translation, all the points in the slider move in the same direction. Thus, the slider's angular velocity remain zero, i.e., $w_z = 0$. In this case, we derive the closed form formula for the friction impulses that acts on the slider during the motion. Furthermore, we derive the equations of pure translation motion. 

The derivation is based on our discrete-time dynamic model (Equations~\eqref{eq:dy_ts_1} to~\eqref{eq:dy_ts_4}). Because $w_z =0$ for each time step, based on Equation~\eqref{eq:dy_ts_3}, the frictional angular impulse $p_r = 0$. Therefore, Equations~\eqref{eq:dy_ts_1},~\eqref{eq:dy_ts_2} and~\eqref{eq:dy_ts_4} can be simplified as:
\begin{align}
\label{eq:p_t_s}
&p_{t}^{u+1}=\frac{-e^{2}_{t}\mu p_{n} (mv_x^{u} + p_x^{u})}{m\sigma^{u+1}+e^2_t\mu p_n}\\
\label{eq:p_o_s}
&p_{o}^{u+1}=\frac{-e^{2}_{o}\mu p_{n} (mv_y^{u} + p_y^{u})}{m\sigma^{u+1}+e^2_o\mu p_n}\\
\label{eq:const_s}
&\mu^2p_n^2 = (p_t^{u+1}/e_t)^2+(p_o^{u+1}/e_o)^2
\end{align}

Then substituting Equations~\eqref{eq:p_t_s} and~\eqref{eq:p_o_s} into~\eqref{eq:const_s}, we get:
\begin{equation}
\frac{(mv_x^{u} + p_x^{u})^2}{(m\sigma^{u+1}+e^2_t\mu p_n)^2}+\frac{(mv_y^{u} + p_y^{u})^2}{(m\sigma^{u+1}+e^2_o\mu p_n)^2} = 1
\end{equation}

Given the isotropic friction assumption, i.e., $e_t = e_o$, we get:
\begin{equation}
m\sigma^{u+1}+e^2_t\mu p_n = \sqrt{(mv_x^{u} + p_x^{u})^2 + (mv_y^{u} + p_y^{u})^2}
\end{equation}

Thus, the analytical solutions for friction impulse are:
\begin{align}
\label{eq:p_t_f}
&p_{t}^{u+1}=\frac{-e^{2}_{t}\mu p_{n} (mv_x^{u} + p_x^{u})}{\sqrt{(mv_x^{u} + p_x^{u})^2 + (mv_y^{u} + p_y^{u})^2}}\\
\label{eq:p_o_f}
&p_{o}^{u+1}=\frac{-e^{2}_{o}\mu p_{n} (mv_y^{u} + p_y^{u})}{\sqrt{(mv_x^{u} + p_x^{u})^2 + (mv_y^{u} + p_y^{u})^2}}
\end{align}

Therefore, we can solve for the velocities from: 
\begin{equation}
v_x^{u+1} =( p_t^{u+1} +p_x^{u})/m+v_x^{u}, \quad
v_y^{u+1} = (p_o^{u+1} +p_y^u)/m+v_y^{u}.
\end{equation}
where $p_t^{u+1}$ and $p_o^{u+1}$ are given by Equations~\eqref{eq:p_t_f} and~\eqref{eq:p_o_f}.


\section{Numerical Results}
In the preceding section, we derived the closed form expressions for computing the velocities of the slider for quasi-static motion and pure translation with isotropic friction assumption. For the general planar sliding motion there does not exist analytical solutions. Therefore, here, we present numerical solutions based on the discrete-time model of quadratic equations that we developed and compare the results with the full nonlinearity complementarity problem (NCP) formulation from~\cite{xie2016rigid}, where we do not assume a priori that the motion is planar.

The first example is of a slider with square contact patch sliding on a horizontal surface. 
We compare solutions from the scheme in this paper to that from~\cite{xie2016rigid} to validate our technique against our previous NCP-based approach, which gives the correct solution. In our second example, we simulate the sliding motion with a ring-shaped contact patch. This example demonstrates that the quadratic model presented in this paper can simulate the sliding motion with wide range of contact shapes (either convex or non-convex). Furthermore, it is not possible to use a few (say four) contact points chosen in an ad-hoc fashion to model the contact patch.  In the third example, we provide a scenario of the slider being pushed with external force. The external force acts on one side of the slider and its position is fixed. The magnitude of external force is periodic and is always perpendicular to  the slider. We use this example to show that the slider with external force can be modeled with our scheme.

We use 'fsolve' in MATLAB, which uses a trust region ('trust-region-dogleg') algorithm to solve the quadratic model. We use PATH complementarity solver~\cite{StevenP.Dirkse1995} to solve the NCP-based model as well as the quadratic model derived in this paper. We compare the average time taken per time-step based on different models for all the examples. The average times per time -step are shown in Table~\ref{table_1}. Since the algorithm in~\cite{xie2016rigid} assumes convex contact patch, the second example could not be solved with this approach. Hence there is no data for this example in Table~\ref{table_1}. As can be seen from the examples, the quadratic model solved in PATH gives consistently better performance. The duration of simulation is shown in paranthesis besides the Example number in Table~\ref{table_1}.  All the examples are implemented in Matlab and run times are obtained on a Macbook Pro with 2.6 GHZ processor and 16 GB RAM.
\comment{
\begin{table}
\begin{tabular}{|l|l|l|}
\hline
Methods  & Example 1 ($0.45s$) \qquad	& Example 2 ($3s$)	\qquad \\ \hline
NCP-based Model (PATH)& $0.0064s$  &$0.0064s$ \\ \hline
Quadratic Model (PATH)& $0.0024s$& $0.0026s$ \\ \hline
Quadratic Model ('fsolve')& $0.0099s$&$0.0083s$  \\ \hline

\end{tabular}
\label{table_1}
\end{table}

}
\begin{table}
\begin{tabular}{|l|l|l|l|}
\hline
Methods  & Example 1 ($0.45s$) \qquad	& Example 2 ($0.6s$)	\qquad & Example 3 ($3s$)\qquad	\\ \hline
NCP-based Model (PATH)& $0.0064s$  &$0.0064s$& $0.0036s$ \\ \hline
Quadratic Model (PATH)& $0.0024s$& $0.0026s$& $0.0014s$ \\ \hline
Quadratic Model ('fsolve')& $0.0099s$&$0.0083s$& $0.0058s$  \\ \hline

\end{tabular}
\caption{Average time taken for each step based on different methods}
\label{table_1}
\end{table}

\label{se:nr}
\begin{figure*}[h]%
\begin{subfigure}{0.49\columnwidth}
\includegraphics[width=\columnwidth]{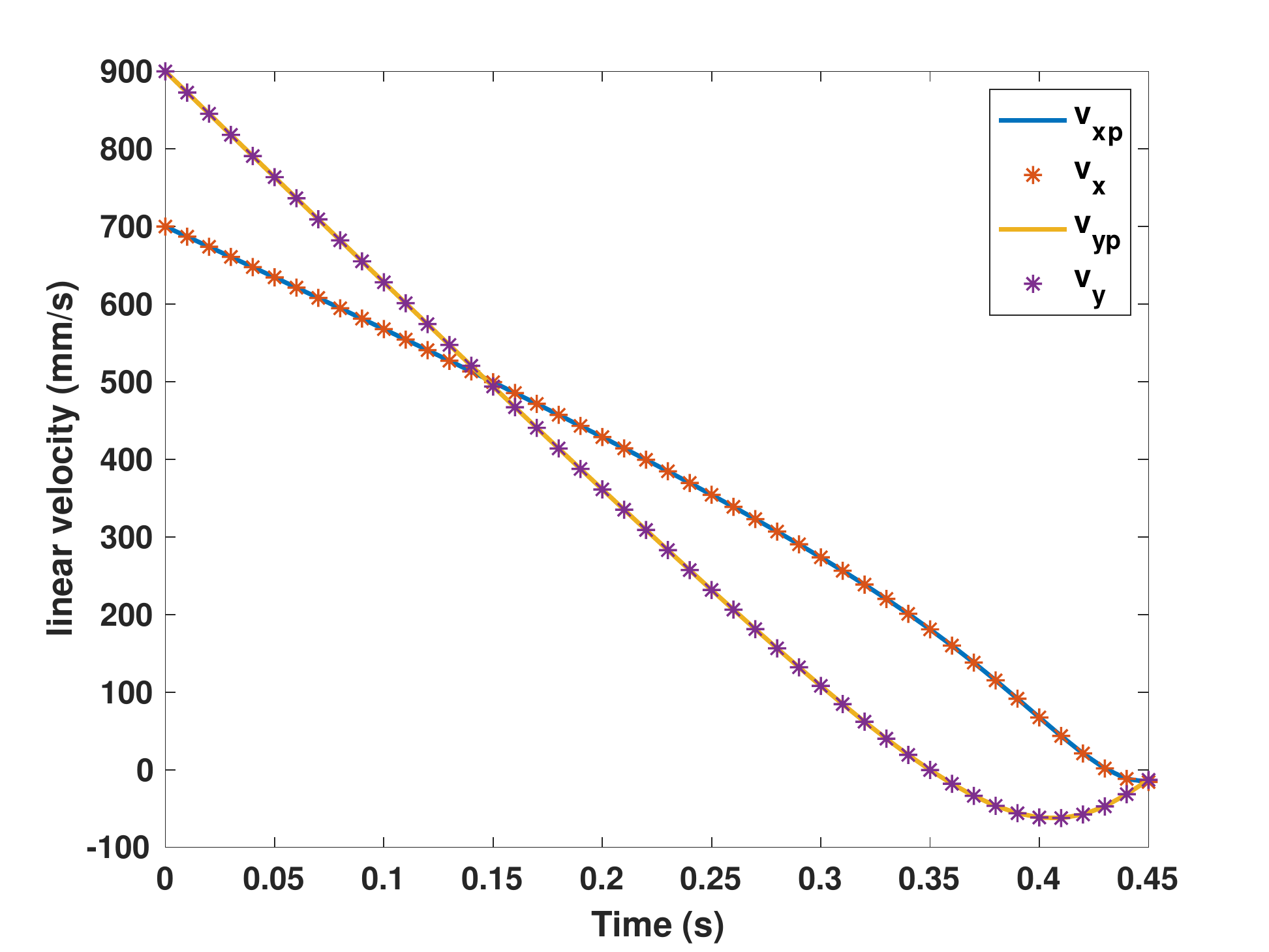}%
\caption{Linear velocity based on our quadratic model ($v_x$ and $v_y$) is same as the solution from our NCP-based model($v_{xp}$ and $v_{yp}$).}
\label{figure:ex1_1} 
\end{subfigure}\hfill%
\begin{subfigure}{0.49\columnwidth}
\includegraphics[width=\columnwidth]{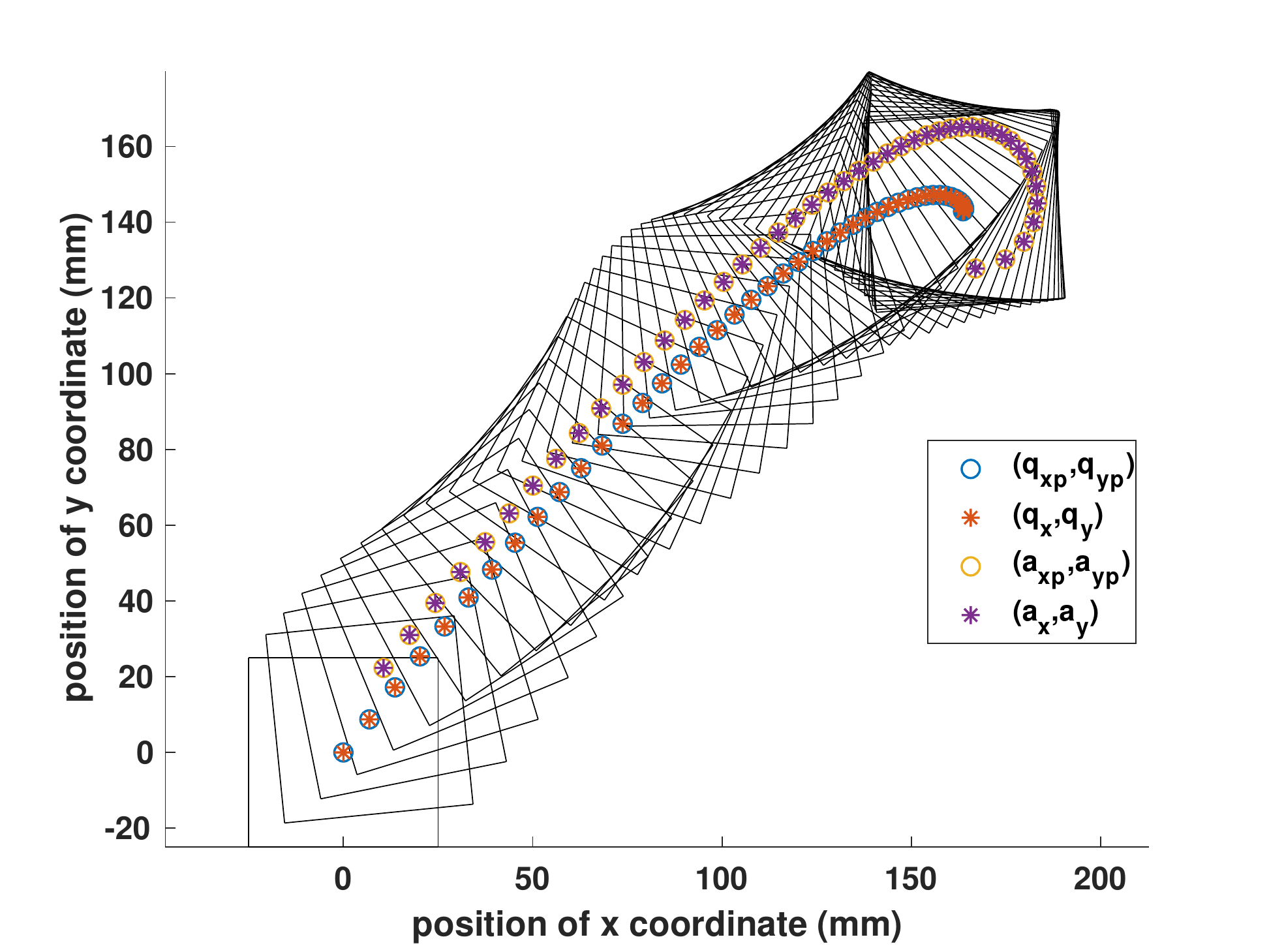}%
\caption{Snapshot of slider's motion. CM and ECP based on discrete-time model ($q_{x},q_{y}$ and $a_x,a_y$) coincide with the result based on NCP model ($q_{xp},q_{yp}$ and $a_{xp},a_{yp}$). }
\label{figure:ex1_3} 
\end{subfigure}%
\caption{Slider with square contact patch slides on the surface without external forces. During motion, the ECP varies within the contact patch. }
\label{Example1}
\end{figure*}

\begin{figure*}[h]%
\begin{subfigure}{0.3\columnwidth}
\includegraphics[width=\columnwidth]{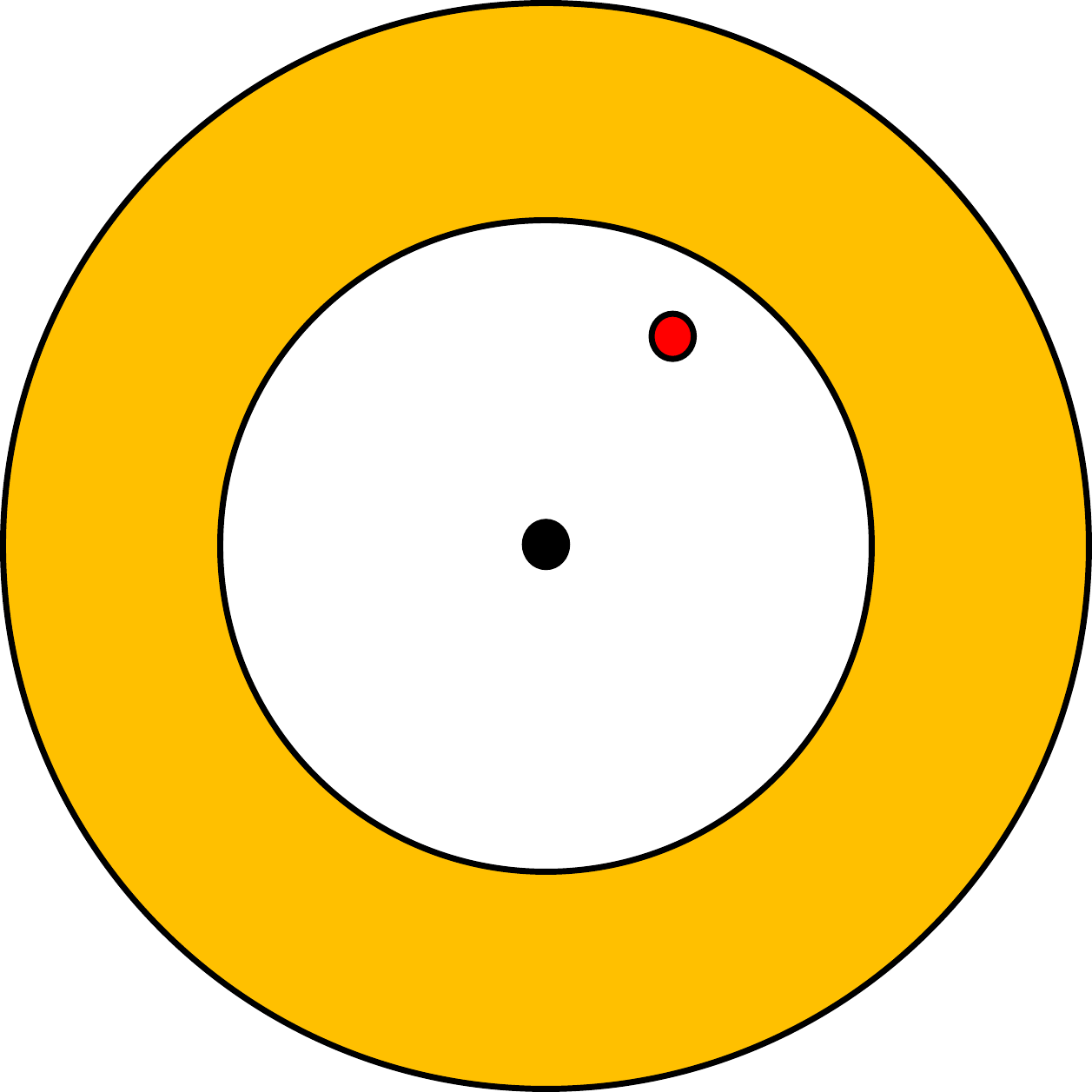}%
\caption{}
\label{figure:ex_non_vonvex_1} 
\end{subfigure}\hfill%
\begin{subfigure}{0.49\columnwidth}
\includegraphics[width=\columnwidth]{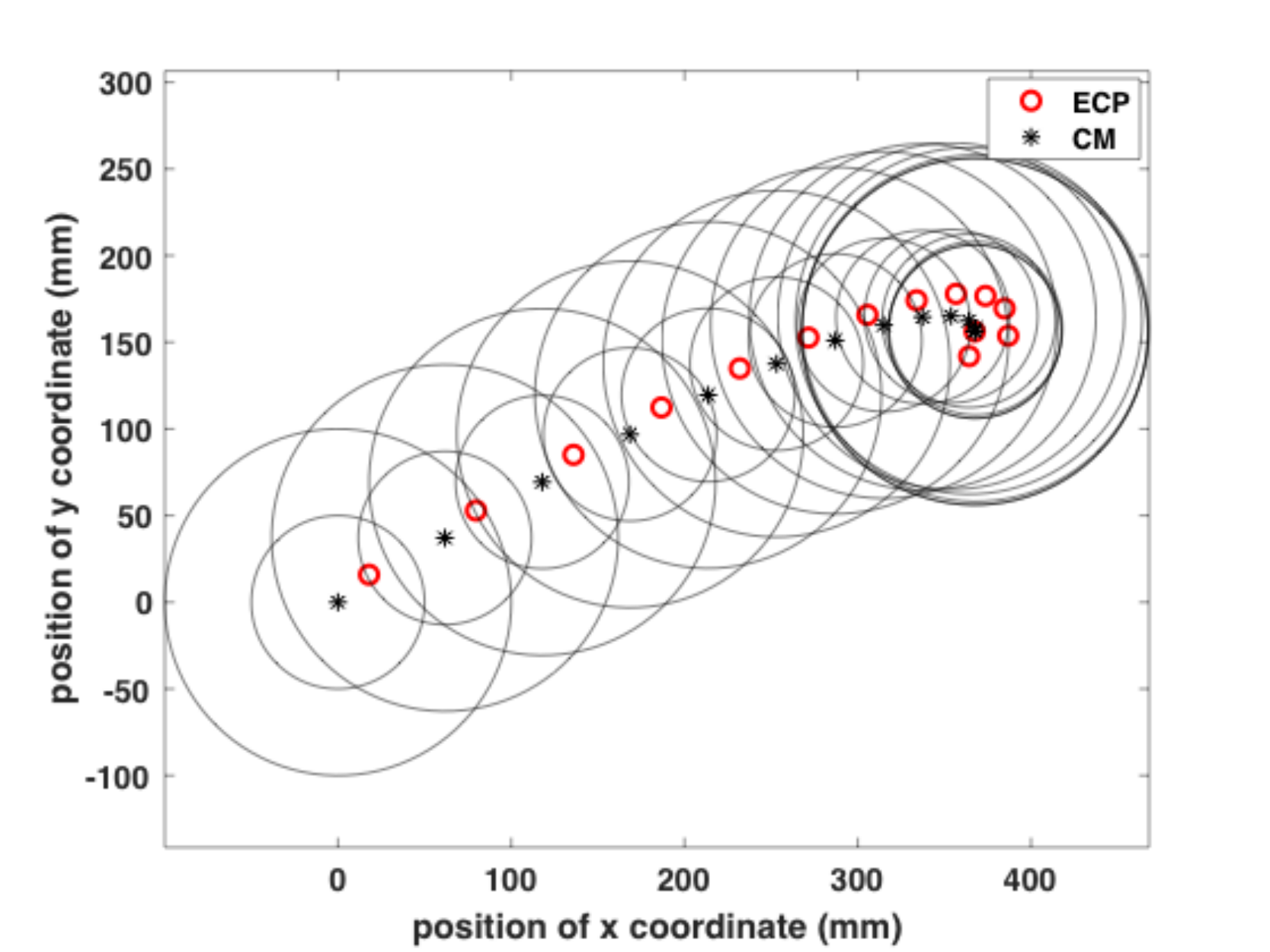}%
\caption{}
\label{figure:ex_non_vonvex_2} 
\end{subfigure}%
\caption{(a) Non-convex contact patch (shaded yellow) between slider and the support. The black dot is the projection of slider's CM on support, and the red dot is one valid position of the ECP, which may not be in the contact patch, but is within the convex hull of the patch. (b) Snapshot of slider's motion based on the quadratic model. Red circle and black star markers represent ECP and CM. During motion, the ECP separates from CM and varies within the contact patch. When it stops, the ECP is just beneath the CM.}
\label{Example1}
\end{figure*}

\begin{figure*}[h]%
\begin{subfigure}{0.5\columnwidth}
\includegraphics[width=\columnwidth]{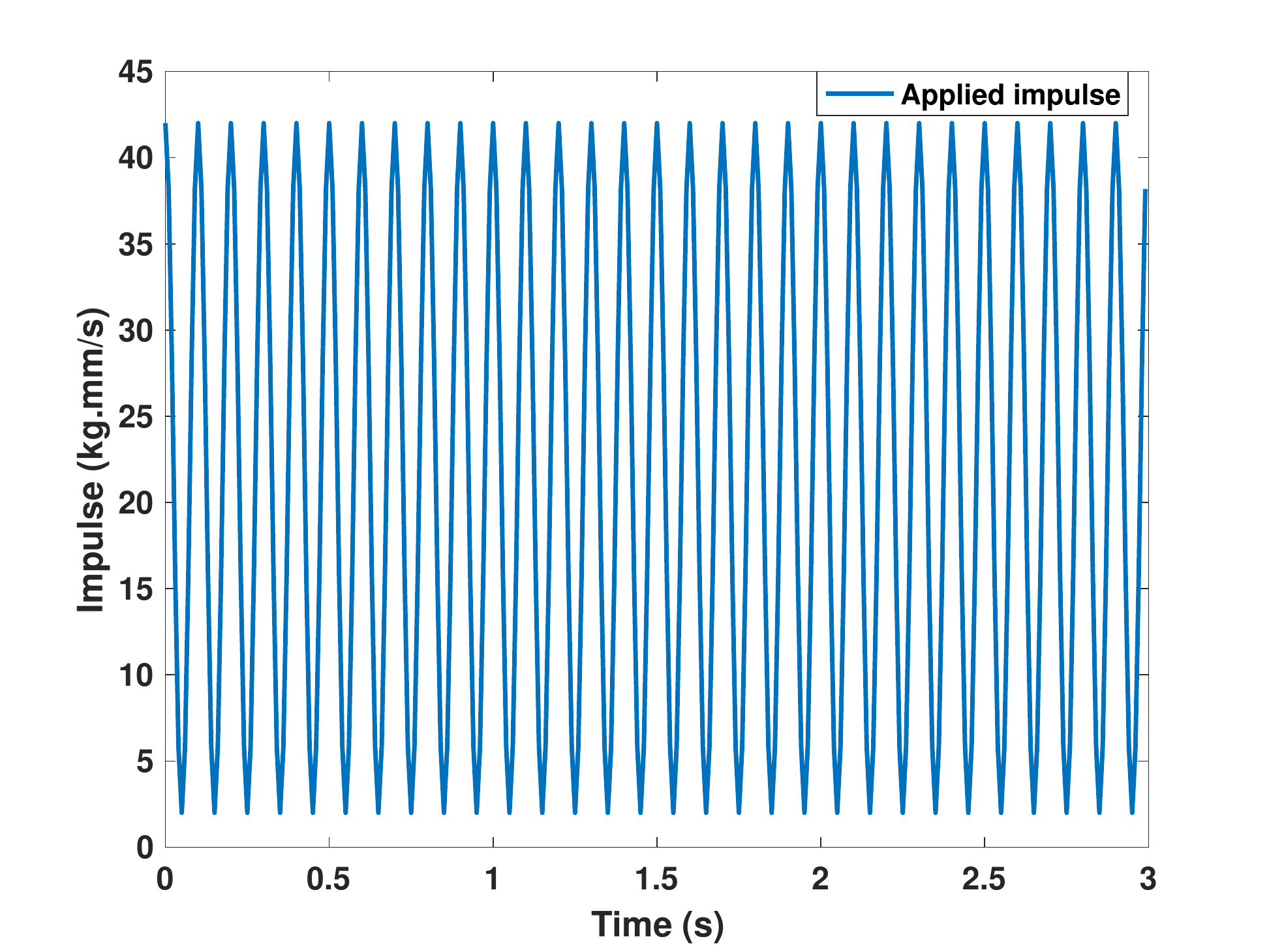}%
\caption{Magnitude of periodical applied impulse.}
\label{figure:ex2_1} 
\end{subfigure}\hfill%
\begin{subfigure}{0.5\columnwidth}
\includegraphics[width=\columnwidth]{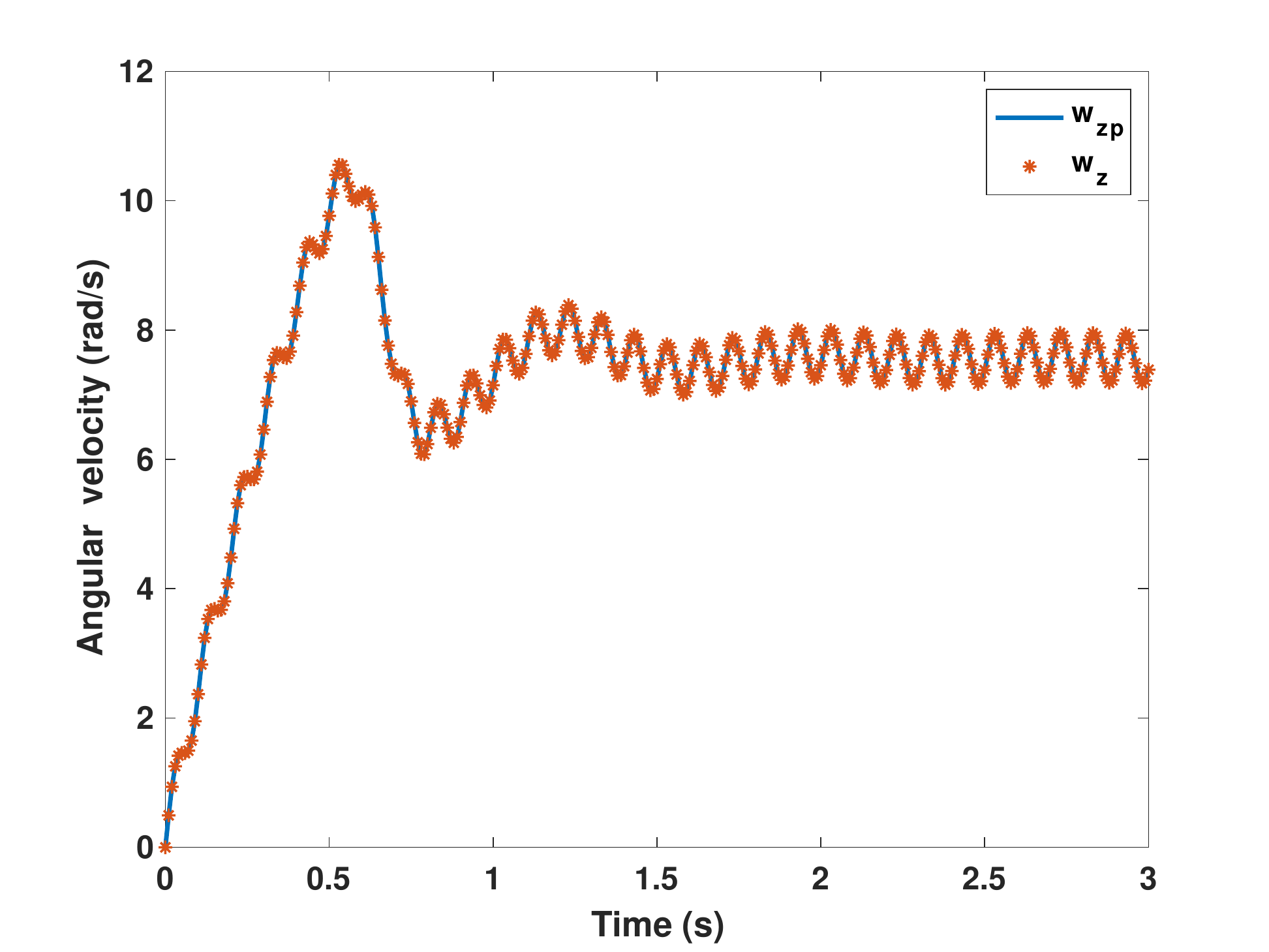}%
\caption{The angular velocity $w_z$ about normal axis.}
\label{figure:ex2_2} 
\end{subfigure}\hfill%
\begin{subfigure}{0.5\columnwidth}
\includegraphics[width=\columnwidth]{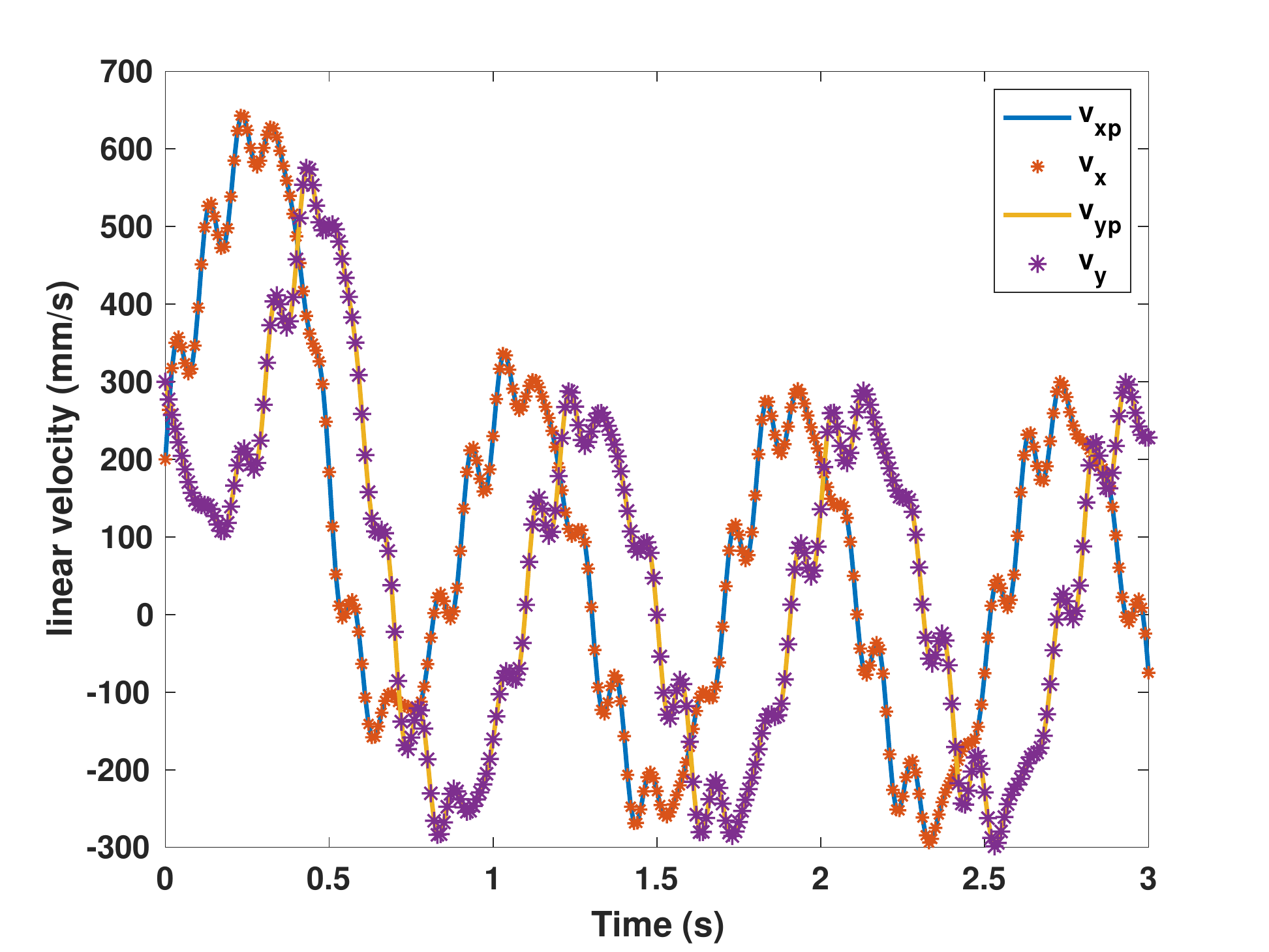}%
\caption{The tangential components of slider's translational velocity.}
\label{figure:ex2_3} 
\end{subfigure}\hfill%
\begin{subfigure}{0.5\columnwidth}
\includegraphics[width=\columnwidth]{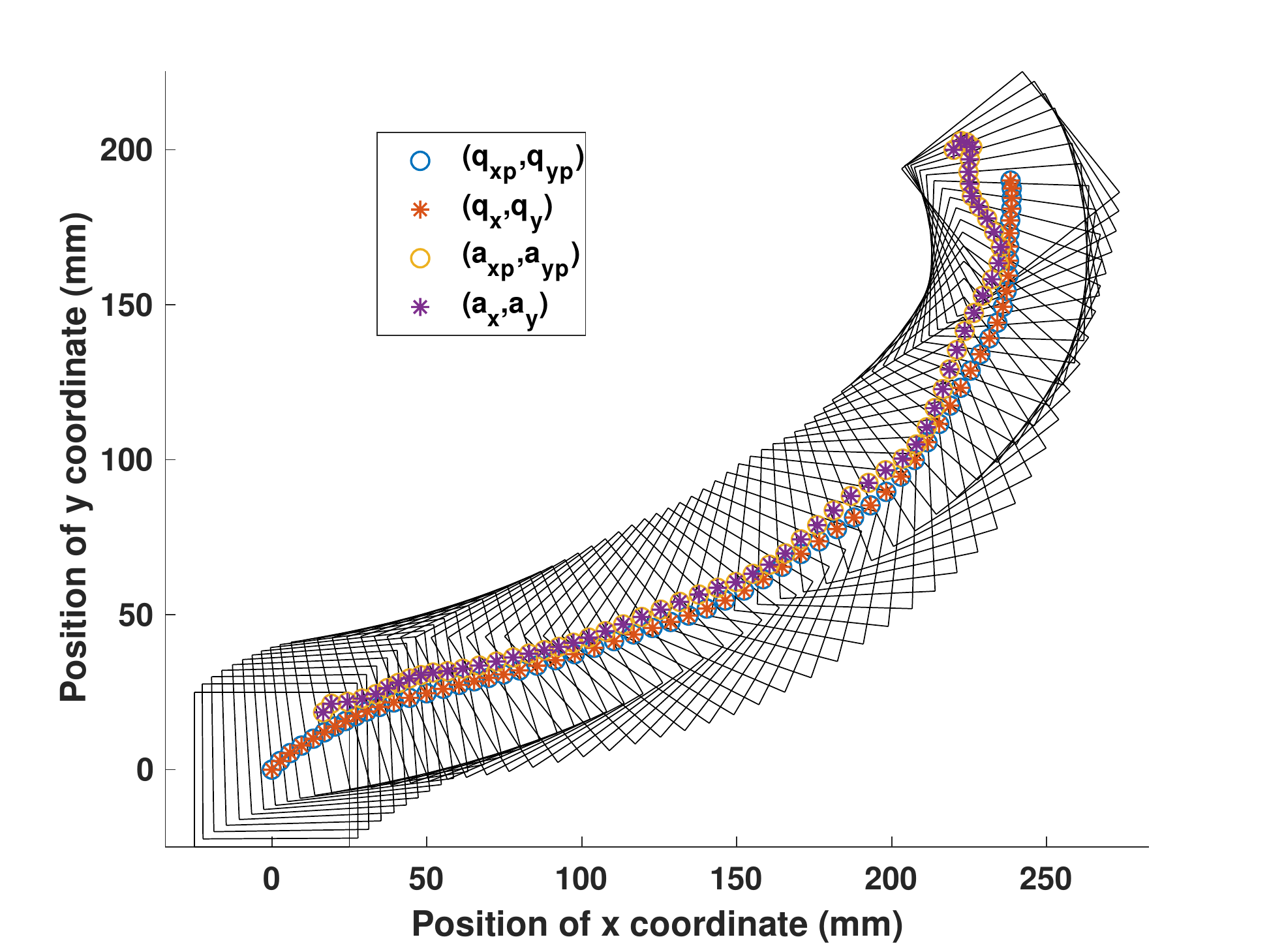}%
\caption{The snapshot of slider's motion from $t= 0s$ to $t = 0.6s$.}
\label{figure:ex2_4} 
\end{subfigure}%
\caption{Slider with square contact patch being pushed by applied force.}
\label{Example2}
\end{figure*}
{\bf Slider on a plane without external forcing:}
In this example, we let a slider with square contact patch slide on the surface. The time step chosen for our discrete-time formula and geometric time-stepping scheme is $h = 0.01s$ and simulation time is $0.45s$. The coefficient of friction between slider and support plane is $\mu = 0.31$, and the friction ellipsoid's given positive constants are: $e_t = e_o = 1$, $e_r = 0.01m$. The mass of the slider is $m = 0.5kg$, and acceleration due to gravity is $g = 9.8m/s^2$. The slider slides on the surface without external force. Its initial position of CM is $q_x = q_y = 0m$, $q_z = 0.08m$. Initial orientation is $\theta = 0^{\circ} $. The initial state of slider is $v_x = 0.7 m/s$, $v_y = 0.9 m/s$ and $w_z = 10 rad/s$. 

As Figure~\ref{Example1} illustrates, a slider slides on the horizontal supporting surface. The forces that act on the slider are friction forces and moments. In Figure~\ref{figure:ex1_1}, we compute the velocities of the slider ($v_x$ and $v_y$) numerically based on our discrete-time dynamic model (Equations~\eqref{eq:dy_ts_1} to~\eqref{eq:dy_ts_4}), and compare it with the result ($v_{xp}$ and $v_{vp}$) from NCP-based model. There exists no difference between two results (within numerical tolerance of $1e-6$ ), which validates our method. In addition, the average time that NCP-based model spends for each time-step is $0.0064s$, our quadratic model's average time is $0.0024s$. This is because our model essentially is a system of $4$ quadratic equations with $4$ variables (there are also a few linear equations afterwards, but the computational cost of those are negligible) , and thus the size and complexity of the system are much less than the NCP-based model (in sliding case, the system is composed of $24$ nonlinear equations and unknowns).
In Figure~\ref{figure:ex1_3}, we plot the snapshots of the slider's contact patch with CM and ECP at each time step. The contact patch's shape is the square with length $L = 0.05m$ and width $W = 0.05m$. During the motion, we observe that the position of ECP always separates from the position of CM, and its relative position to the slider frame $\mathcal{F}_s$ changes within the contact patch. The observation confirms that when the position of center of mass is above the support plane, acceleration of object would cause the shift of ECP~\cite{mason2001mechanics}. 

{\bf Sliding motion with ring-shaped contact patch:}
In this example, the slider has ring-shaped contact patch with the support. As Figure~\ref{figure:ex_non_vonvex_1} illustrates, the contact patch is the ring in yellow, which is non-convex and can not be represented as the convex hull of three chosen support points. If the ECP or center of friction is outside the convex hull, then the motion predicted would be inaccurate. 

We use the quadratic model presented in this paper to simulate the motion of the slider. The time step and friction parameters chosen are same as in first example. The mass of the slider would be $m = 1kg$. The initial configuration of the slider is also same as in first example. The initial state is chosen to be $v_x = 1.3 m/s, v_y = 0.8 m/s, w_z = 11 rad/s$. The total simulation time is $0.65s$.
In Figure~\ref{figure:ex_non_vonvex_2}, we plot the snapshot of the ring-shaped contact patch with ECP and CM at each time step. The radius for the outer circle is $0.1m$, while the radius for the inner circle is $0.05m$. During the sliding motion, the ECP always separates from CM. When the slider stops, the ECP is just beneath the CM.

{\bf Slider being pushed on a plane:}
In this example, we let the slider be pushed by applied force on the horizontal plane. The dimension and mass of the slider as well as the friction parameters ($\mu$, $e_t$, $e_o$ and $e_r$) is same as previous example. The initial state of slider to be $v_x = 0.2 m/s$, $v_y = 0.3 m/s$ and $w_z = 0 rad/s$. The time-step is still $0.01s$ and simulation time is extended to $3s$. During the motion,  as shown in Figure~\ref{figure:ex2_4}, we apply the force on the left edge of the square patch. The position of external force on the edge is fixed at $2.5mm$ below the middle of the edge. Let applied force always be perpendicular to that side during the motion. As Figure~\ref{figure:ex2_1} illustrates, the magnitude of the applied force is periodic, i.e., $F_{push} = 2.2 + 2\cos(2\pi t/T) \ N $, where the period $T = 0.1s$.

From Figure~\ref{figure:ex2_2}, we compute the angular velocity about normal axis $w_z$ based on our quadratic model and $w_{zp}$ based on our NCP-based model. For Figure~\ref{figure:ex2_3}, we compute the linear components of velocity $v_x$, $v_y$ based on our quadratic model, and $v_{xp}$, $v_{xp}$ based on our NCP-based model. For both angular and tangential velocities, the difference between quadratic and NCP solutions is within the numerical tolerance of $1e-6$. Figure~\ref{figure:ex2_4} plot the snapshot of the slider at each time step between the time period from $t = 0s$ to $t = 0.6s$.

\section{Conclusions}
In this paper, we present a quadratic discrete-time dynamic model for solving the problem of general planar sliding with distributed convex contact patch. Previous method assumes quasi-static motion or chooses multiple contact points (usually three) in an ad-hoc manner to approximate the entire contact patch. In our dynamic model, the effect of contact patch is equivalently modeled as the contact wrench at the {\em equivalent contact point}. During the motion, the balance of all the external forces and moments including gravity force, applied force and frictional force fixes the position of ECP. Therefore, by combing the equation of motion with friction model, we get the quadratic discrete-time model. This allows us to solve two components of tangential friction impulses, the friction moment and the slip speed. The state of the slider as well as the ECP can be computed by solving a system of linear equations once the contact impulses are computed. In addition, we also provide closed form expression for quasi-static motion and pure translation motion. We also demonstrate the numerical results based on our quadratic model and NCP model for the general planar motion of the slider with or without applied force. In the appendix section, we provide the closed-form expressions for the friction parameters ($\mu, e_t, e_o, e_r$) based on our quadratic discrete-time dynamic model. In the future, we would like to use the expressions developed for estimating the contact parameters. 
%
%
\bibliographystyle{IEEEtran}

\section*{Appendix A: System Identification}
In this section, we show that the equations of motion derived for sliding motion, i.e., Equations~\eqref{eq:dy_ts_1} to~\eqref{eq:dy_ts_4} can be used for identifying the parameters of the friction model.
The system identification problem that we consider is as follows: {\em Given the mass and moment of inertia of the slider (i.e., $m$ and $I_z$), the discrete-time trajectory of the slider, i.e., $v_x^{u}, v_y^{u}, \omega_z^{u}, \forall u$, and the history of applied wrenches, i.e., $p_x^{u}$, $p_y^{u}$, $p_{x\tau}^u$, $p_{y\tau}^u$, $p_{z\tau}^u$, $\forall u$, compute the friction parameters $\mu$, $e_t$, $e_o$, and $e_r$}. 

We will now derive closed-form expressions for the friction parameters ($\mu$, $e_t$, $e_o$, $e_r$) based on the one-step discrete-time dynamic model (Equations~\eqref{eq:dy_ts_1} to~\eqref{eq:dy_ts_4}). From Equations~\eqref{eq:dy_ts_1} to~\eqref{eq:dy_ts_4}, by algebraic simplification, we obtain
\begin{align}
\label{SI:1}
&e_t^2\mu = \left(\frac{p_t^{u+1}}{p_n}\right)^2 + \frac{p_t^{u+1}p_o^{u+1}v_o^{u+1}}{p_n^2v_t^{u+1}}+\frac{p_t^{u+1}p_r^{u+1}v_r^{u+1}}{p_n^2v_t^{u+1}} \\
\label{SI:2}
&\left(\frac{e_o}{e_t}\right)^2 = \frac{p_o^{u+1}v_t^{u+1}}{p_t^{u+1}v_o^{u+1}}; \qquad
\left(\frac{e_r}{e_t}\right)^2 = \frac{p_r^{u+1}v_t^{u+1}}{p_t^{u+1}v_r^{u+1}}
\end{align}
where friction impulses $p_t^{u+1}$, $p_o^{u+1}$, $p_r^{u+1}$ and velocity components $v_t^{u+1}$, $v_o^{u+1}$, $v_r^{u+1}$ are computable based on trajectory of the slider and history of applied wrenches.
\begin{equation}
\begin{aligned}
&p_t^{u+1} = m(v_x^{u+1} - v_x^u)-p_x^u \\ &p_o^{u+1} = m(v_y^{u+1} - v_y^u)-p_y^u\\ &p_r^{u+1} = I_z(w_z^{u+1} - w_z^u)-p_{z\tau}^u\\
&v_t^{u+1} = v_x^{u+1} -w_z^{u+1}(-p_{x\tau}^{u}-p_o^{u+1}q_z)/p_n\\ &v_o^{u+1} = v_y^{u+1} +w_z^{u+1}(p_{y\tau}^{u}-p_t^{u+1}q_z)/p_n\\ &v_r^{u+1} = w_z^{u+1}
\end{aligned}
\end{equation}

Note that Equations~\ref{SI:1} and~\ref{SI:2} provide us the closed-form expressions for $e_t\mu$ (the coefficient of friction along $\bm{t}$ axis of contact frame), $\frac{e_o}{e_t}$ and $\frac{e_r}{e_t}$. Based on Equation~\ref{equation:friction}, parameters $e_t, e_o, e_r, \mu$ would be redundant to determine the friction ellipsoid. Thus, we choose $e_t\mu, \frac{e_o}{e_t}, \frac{e_r}{e_t}$ as our friction parameters, which would be sufficient to determine the friction ellipsoid.

If the friction parameters are constant and the measurements noiseless, then the one-step estimate of the friction parameters should be the same across all time-steps and the estimate from just one step allows us to identify the parameters. However, in practice, the measurement of velocity of the slider as well as the applied wrenches are noisy. Therefore, we have to use statistical estimation techniques to obtain estimate of the contact parameters. The development of the estimation techniques is left as future work. However, the fact that the friction parameters can be expressed in closed form makes the estimation problem easier.

\end{document}